\pdfoutput=1

\documentclass[11pt]{article}
\usepackage{graphicx}
\graphicspath{./images/} 
\usepackage{lscape}
\usepackage{amsmath}
\usepackage{amssymb}
\usepackage{EMNLP2022}

\usepackage{times}
\usepackage{latexsym}

\usepackage[T1]{fontenc}

\usepackage[utf8]{inputenc}

\usepackage{microtype}

\usepackage{inconsolata}

\title{Scientific and Creative Analogies in Pretrained Language Models}

\author{Tamara Czinczoll$^{\clubsuit\heartsuit}$ ~ Helen Yannakoudakis$^{\spadesuit}$ ~ Pushkar Mishra$^{\diamondsuit}$ ~ Ekaterina Shutova$^{\clubsuit}$ \\
$^\clubsuit$ILLC, University of Amsterdam, the Netherlands \\
$^\diamondsuit$Meta AI, London, United Kingdom \\
$^\spadesuit$Dept. of Informatics, King's College London, United Kingdom\\
$^\heartsuit$Hasso Plattner Institute/University of Potsdam, Germany\\
{ \small \tt tamara.czinczoll@hpi.de, helen.yannakoudakis@kcl.ac.uk, pushkarmishra@meta.com, e.shutova@uva.nl}}

\begin{document}
\maketitle
\begin{abstract}

This paper examines the encoding of analogy in large-scale pretrained language models, such as BERT and GPT-2.
Existing analogy datasets typically focus on a limited set of analogical relations, with a high similarity of the two domains between which the analogy holds. As a more realistic setup, we introduce the \textbf{S}cientific and \textbf{C}reative \textbf{An}alogy dataset (SCAN), a novel analogy dataset containing systematic mappings of multiple attributes and relational structures across dissimilar domains. Using this dataset, we test the analogical reasoning capabilities of several widely-used pretrained language models (LMs). 
We find that state-of-the-art LMs achieve low performance on these complex analogy tasks, highlighting the challenges still posed by analogy understanding.
\end{abstract}

\section{Introduction}

Analogy-making is a cornerstone of human intelligence \citep{gentner2001}, allowing us to acquire new knowledge and creatively explore new concepts. 
According to \citet{gentner1983}'s \emph{Structure-Mapping Theory}, analogical reasoning is different from surface similarity. Instead, the attributes of a familiar concept (the source domain) are mapped to something less familiar (the target domain) if their \emph{relational structures} are similar enough. For example, while not directly similar in their attributes, the underlying relational structure of the \textit{solar system} matches that of an \textit{atom}. The relationship between two source domain attributes (e.g. \emph{sun} and \emph{planet}) helps us to understand that between their target domain counterparts (\emph{nucleus} and \emph{electron}).

Within natural language processing (NLP), the word analogy task \citep{mikolov2013}, has been widely used to demonstrate analogical reasoning capabilities of pretrained word embedding models. The task involves solving analogies of the form \emph{A:B :: C:D} (i.e., \emph{A is to B as C is to D}) by exploiting (local) linear properties of word vectors (vector offsets).
Subsequently, word analogy became one of the standardized tasks for intrinsic evaluation of word embedding quality. However, \citet{GladkovaDrozd2016} showed that the vector offset method was not sufficient for most types of analogical relations, and \citet{rogers2017analogy} pointed out shortcuts that the models were taking.
Existing word analogy datasets focus on a limited set of analogical relations, include words that are semantically similar and do not require the model to relate distinct concepts via systematic comparison of their relational structures, all of which is necessary for human-like analogy making.
In parallel, the field has seen the development of large-scale pretrained sentence encoders, whose analogical reasoning capabilities have not yet been fully tested.

To address these issues, we devise and release a new dataset -- the \textbf{S}cientific and \textbf{C}reative \textbf{An}alogy dataset (\textbf{SCAN}) -- comprising holistic analogies between concepts from semantically distant domains. It draws on metaphorical and scientific analogies. Resolving these analogies requires the models to identify systematic ontological correspondences between two distinct semantic domains, such as in the \emph{solar system -- atom} example. 
Our contributions are threefold:
1) We present the \textbf{SCAN} analogy evaluation task and dataset, which we make publicly available to the research community;
2) We systematically evaluate current state-of-the-art LMs on the established BATS dataset \citep{GladkovaDrozd2016}, which consists of a large number of traditional word analogies, as well as the novel SCAN dataset. We show that in the latter case the models exhibit severe limitations in understanding analogies;
3) We show that a high performance on BATS is not indicative of how well the models solve the complex SCAN analogies, supporting our hypothesis that BATS does not require full analogical reasoning.

\section{Related Work}

\citet{Turney2008} presented an algorithm for analogy solving and tested it on 20 scientific and metaphorical examples, where a source domain is mapped to a different target domain, along with a number of its attributes (e.g. \emph{waves} to \emph{sound}). While few, these examples were true to human analogy-making, representing a wide range of semantic relationships.
Using a linear offset method (\emph{3CosAdd}), \citet{mikolov2013} demonstrated that their word embeddings automatically capture analogical information about word relationships, so that $emb(king) - emb(man) + emb(woman) \approx emb(queen)$, where $emb$ is the embedding function represented by the neural network. \citet{GladkovaDrozd2016}, however, showed that the pretrained word embeddings at the time could only reliably complete word analogies for inflectional morphology categories while struggling on many semantic categories. They released a balanced, larger and more diverse dataset than \citet{mikolov2013}'s (40 vs 15 relation types), the \emph{Bigger Analogy Test Set} (BATS), demonstrating that Word2Vec was not able to solve most types of word analogies. In particular, a larger semantic distance between the source and target domains resulted in low performance \citep{rogers2017analogy}. 

Transformer language models such as BERT \citep{devlin2019bert} have pushed the state-of-the-art on a number of NLP tasks. But since 3CosAdd cannot easily be applied to them, due to their word embeddings not being fixed but dynamically calculated, their analogical capabilities have not been investigated much.
While some headway has been made in that regard (\citet{li2020learning}, \citet{zhu-de-melo-2020-sentence}, \citet{ushio-2021-bert}), the focus has been on transferring the traditional word analogy datasets to the sentence level. This does, however, also transfer their limitations.
In general, the low performance of models in \citet{GladkovaDrozd2016} and \citet{zhu-de-melo-2020-sentence} suggest that completing word analogies is challenging for state-of-the-art LMs, even when structure mapping across distinct domains is not explicitly tested.

\section{SCAN Dataset}

Our dataset contains 449 analogy instances, clustered into $65$ full concept mappings. A source concept is mapped to a target concept along with a number of related attributes. Table \ref{tab:dataset-examples} provides two examples. When mapping from the source concept \emph{War} to the target concept \emph{Argument}, a number of relevant attributes' correspondences are given. The number of attributes per cross-domain mapping is not fixed. 
The dataset includes the $20$ mappings from \citet{Turney2008} ($10$ scientific and $10$ metaphorical) and extends them by another $43$ metaphorical mappings and $2$ scientific ones. The new metaphorical mappings include conceptual metaphors from the Master Metaphor List \cite{MastMetList} and other conceptual metaphors widely-discussed in linguistic literature \citep{LakoffAndJohnson,Musolff,LakoffWehling}. Each conceptual metaphor was then annotated for attribute correspondences by three metaphor experts. First, the semantic frames of the source and target domains were identified and then the correspondences between individual frame elements were established (see Tab. \ref{tab:dataset-examples}). We build a word analogy task from this data by defining the cross-domain mappings (e.g., \emph{Argument} and \emph{War}) as the first word pair, and the attribute mappings (in this case \emph{Debater} and \emph{Combatant}) as the to-be-completed second word pair. Since each concept includes multiple attributes, a total of $449$ word analogies are constructed.

SCAN offers richer and more holistic analogies than traditional word analogy datasets. Taking a statistical view, the chances of the words in the source and target domains co-occurring are much lower than in BATS. For example, countries and their capitals, animals and the sounds they make, and most grammatical analogy types in BATS are quite likely to occur in the training corpus together. However, the same cannot be said for most SCAN analogies, meaning that true analogical transfer needs to occur.

Additionally, the in-domain words in SCAN are semantically more dissimilar. For example, in the argument domain, \emph{debater} and \emph{topic} are fundamentally different concepts. In BATS, on the other hand, every domain member is another instance of the same concept, e.g. \emph{France} and \emph{Germany} instances of a country.

Lastly, the analogical relationships between domains in SCAN are more abstract than those in BATS. To successfully extract the same relationship from \emph{solar system -- atom} and \emph{planet -- electron}, more abstraction and inference over the relational structure of the domains is needed than in BATS. In BATS, the relationships between, e.g. \emph{France -- Paris} and \emph{Germany -- Berlin}, are straightforward and do not require much abstraction.

Overall, SCAN offers more human-like analogies by employing more diverse in-domain words, more abstract mapping relations and by avoiding obvious co-occurrences. Due to its full-concept mappings, SCAN is not confined to the word analogy task. By holistically mapping entire source domains to a new target domain we want to encourage a broader range of analogy representations.

\begin{table}
\centering
\begin{tabular}{ll|ll}

\textbf{Target} & \textbf{Source} & \textbf{Attribute} & \textbf{mapping}\\
\hline
Argu- &War &Debater &Combatant \\
ment & &Topic &Battleground\\
 &  &Claim &Position\\
 &	&Criticize &Attack \\
 & &Rhetoric &Maneuver\\
\hline
Code &Virus &Malware &Virus \\
 & &Replication &Reproduction \\
 & &Installation &Infection\\
 & &Removal &Eradication\\
 & &Antivirus &Vaccine\\
\end{tabular}
\caption{Example mappings in SCAN. For one source concept multiple relevant attributes are mapped to the corresponding target concept's attributes.}
\label{tab:dataset-examples}
\end{table}

\section{Models}
We probe the analogical capabilities of several widely used language models: GPT-2, BERT and Multilingual BERT (M-BERT). 
We use GloVe as a baseline, given it has been shown to outperform language models on some relation types in previous analogy tasks \citep{zhu-de-melo-2020-sentence}. 

\paragraph{GPT-2} \citep{Radford2019LanguageMA} can be viewed as a ``true'' LM since it is trained to predict the next word in a sequence, and can be used for language generation. It is a transformer-based model with $48$ layers and $1542$M parameters, trained on a custom dataset, \emph{WebText}, created only from outbound links from Reddit to improve text quality. Due to its predictive nature, GPT-2 is one-directional, i.e. only the context on the left-hand side
influences the prediction of the next word.

\paragraph{BERT} \citep{devlin2019bert} is a bidirectional language representation model. It is trained with two objectives: masked-token prediction and next-sentence prediction.
We use BERT-base with 12 layers in our experiments ($110$M parameters). Since BERT is bidirectional, it can incorporate information from both sides of the masked token.

\paragraph{M-BERT} is a BERT model, trained on a Wikipedia dump of $100$ languages.
The model performs best on high-resource languages such as English, French and Chinese, since lower-resource languages are underrepresented in the training data.
We test whether M-BERT's pre-training on a wide range of languages, and thus a wide range of culture-specific analogies, might enhance the model's general analogy understanding.

\section{Experiments}

\paragraph{Setup}

We use pretrained model instances of GPT-2, BERT Base and Multilingual BERT \footnote{\url{https://huggingface.co/}}.
As BERT and GPT-2 are trained on full sentences, we insert the word analogy quadruple into a placeholder sentence. We use ``\textit{If} A \textit{is like} B, \textit{then} C \textit{is like} D.'', which was selected from a set of candidates as it performed best on the development set. Similarly to \citet{ettinger-2020-bert}, who probed BERT with a number of cloze and negation tasks, the models need to predict the last token of the sentence. We force the models to predict word D by either masking it for the two BERT models, or by cutting the sentence off before it for GPT-2.  We report the mean reciprocal rank (MRR) of the first token of the target word (or that of one of the alternative answers) in only the top $10$ predicted tokens to reduce compute. If the label is not in the top $10$ tokens, its RR is $0$. Model performance in terms of accuracy, recall@10 and recall@5 is reported in the Appendix. We use an Nvidia 16GB GPU.

\paragraph{SCAN vs. BATS}
To evaluate how well the models can solve different types of analogy, we test them on \textbf{BATS} in addition to SCAN. We do not fine-tune the models. BATS consists of $98000$ examples of balanced relations. There are four main relations -- inflectional and derivational morphology, and lexicographic and encyclopedic semantics -- each of which consists of ten subcategories. For some examples, multiple correct answers are listed.

\paragraph{Zero-shot vs. One-shot}
Previous work on analogical reasoning in GPT-3 \citep{analogygpt3, analogyblogfantasy} has shown that when the model is given a full example of a word analogy in addition to the incomplete one, the performance on the incomplete analogy substantially increases. We see this as a form of one-shot vs. zero-shot testing and also test the models this way, investigating whether this has an impact on the LMs' performance on SCAN. We insert a complete version of our template sentence before the incomplete one, ensuring that none of the analogy words from the full example appear in the incomplete analogy. Note that GloVe does not benefit from this setup as the vectors used for 3CosAdd remain the same.

\paragraph{Training Set Effects}
Lastly, we further investigate the difference between the  types of word analogies in BATS and those in SCAN. We split the BATS dataset into a train, validation and test set (70/15/15 ratio), ensuring that each word pair appears in only one of them. We fine-tune the LMs on the training set and take each model's version with the best score on the BATS validation set. We train ($\sim4$h) all models with the AdamW \citep{loshchilov2019adamw} optimizer, a learning rate of $5e^{-5}$ and a batch size of $16$ for $4$ epochs (based on manual tuning).
If the model has learned about general analogy-making it must understand new analogical relations “on-the-fly” and improve not only on the BATS test set but also on SCAN. We expect there to be strong improvements on BATS compared to its untrained counterpart, but little to none on SCAN, showing them to be inherently different. This is not performed on GloVe, since 3CosAdd outputting an embedding is not part of its original training setup.

\section{Results \& Discussion}

Table \ref{tab:results-zero-shot} shows the accuracy of each of the models on the BATS and SCAN datasets, as well as on the scientific and creative analogies separately. 
\begin{table}
\centering
\scalebox{0.8}{
\begin{tabular}{lcccc}
\hline
 & \textbf{BATS} &\textbf{SCAN} &\textbf{Science} & \textbf{Meta.}\\
\hline
\textbf{GloVe} &.099 &.022 &.099 &.006\\
\textbf{GPT-2} &.098 &\textbf{.057} &.073 &\textbf{.054}\\
\textbf{BERT} &\textbf{.207} &\textbf{.044} &.092 &\textbf{.034}\\
\textbf{M-BERT} &\textbf{.205} &.041 &.088 &\textbf{.031}\\
\hline
\end{tabular}}
\caption{Model MRR on BATS and SCAN. Statistically significant differences compared to the GloVe baseline are in bold (two-sided permutation test; $p<0.05$; \#resamples=$10e^5$).}
\label{tab:results-zero-shot}
\end{table}
BERT achieves the highest MRR on the BATS dataset, with a strong lead compared to the other models. Similarly to \citet{zhu-de-melo-2020-sentence}, we find that GloVe can keep up with the other models on BATS, performing similarly to GPT-2. However, this trend is not observed on the SCAN dataset, where GloVe is relegated to last place, indicating that context is important for understanding SCAN analogies. All models perform better on the scientific analogies than on metaphors. This could be due to the fact that their attributes are less abstract and have clearer correspondences in the target domain. The models' MRRs are generally lower on SCAN, which we attribute to the greater semantic dissimilarity between source and target domains. GPT-2 achieves the highest performance, followed by BERT. This, combined with its lower accuracy on the BATS baseline, indicates that GPT-2 is better at modeling more extensive and narrative analogies instead of the more artificial and strictly-defined ones in BATS. Multilingual features only appear to be marginally effective for the task, something which can be explained by the fact that most analogies are language-dependent, an observation also made by \citet{ulcar-etal-2020-multilingual}. Overall, these results indicate that the SCAN analogy task is challenging for state-of-the-art LMs and that their true analogy solving capabilities still need to be improved.

\paragraph{Zero vs. One-Shot}
\begin{table}
\centering
\scalebox{0.8}{
\begin{tabular}{lcccc}
\hline
 & \textbf{BATS} &\textbf{SCAN} &\textbf{Science} & \textbf{Meta.}\\
\hline
\textbf{GPT-2} &\textbf{.121} &.048 &.056 &.046\\
\textbf{BERT} &\textbf{.095} &.035 &.077 &.027\\
\textbf{M-BERT} &\textbf{.180} &.036 &.112 &.020 \\
\hline
\end{tabular}}
\caption{MRR when an example sentence is given. Statistically significant differences (two-sided permutation test; $p<0.05$; \#resamples=$10e^5$) compared to each model's baseline in bold.}
\label{tab:results-one-shot}
\end{table}
Table~\ref{tab:results-one-shot} shows model accuracy when the input contains a complete additional example. Apart from GPT-2 on BATS, this does not help the models better understand the task. This contrasts the examples on GPT-3 from \citet{analogygpt3}, possibly due to the models not identifying the analogical relationships in the example sentence.

\paragraph{Training Set Effects}
\begin{table}
\centering
\scalebox{0.8}{
\begin{tabular}{lcccc}
\hline
 & \textbf{BATS} & \textbf{SCAN} &\textbf{Science} &\textbf{Meta.}\\
\hline
\textbf{GPT-2} &\textbf{.384} &\textbf{.022} &.066 &\textbf{.012}\\
\textbf{BERT} &\textbf{.592} &\textbf{.019} &.061 &\textbf{.010}\\
\textbf{M-BERT} &\textbf{.499} &\textbf{.020} &.087 &\textbf{.006} \\
\hline
\end{tabular}}
\caption{MRR on the BATS test set as well as on SCAN after training. Statistical significance  compared to the baseline (two-sided permutation test; $p<0.05$; \#resamples=$10e^5$) in bold.}
\label{tab:results-training}
\end{table}
After training on BATS, one could expect that if the models learn about analogical reasoning in general, they would also naturally do better on the SCAN dataset with more complex analogies. However, our results in Table~\ref{tab:results-training} show that the opposite is the case. While training on BATS drastically increases the models' MRR on the held-out BATS test set, it has an adverse effect on SCAN. This suggests that the two datasets' analogy types are innately different, validating our hypothesis that standard word analogy datasets do not adequately represent human analogy use.

\paragraph{Error Analysis}
While GloVe scores consistently on all relation types in BATS, this is not the case for the other models. On SCAN, none of the models predict the mappings of all attributes of a concept (or even most of them) correctly. While the models are able to solve some individual mappings, the fact that they cannot apply this to all aspects of the concept indicates that none of them are really able to grasp the workings of analogy. In cases where analogies remain entirely unsolved, it is likely that the required domain knowledge is lacking. 

\section{Conclusion}
Analogical reasoning remains a challenging task even when state-of-the-art Transformer LMs are used. We have shown that, even with models such as BERT and GPT-2, there is large room for improvement on automated reasoning and understanding of realistic analogies. 
We have introduced a new dataset, SCAN, that is different from existing word analogy datasets in that it is composed of whole concept mappings across semantically dissimilar domains, demonstrating that popular LMs are unable to fully understand these analogies. We further tested whether a full example of the task can help the models, finding that this is not helpful in our setup. Lastly, our results indicate that the SCAN analogies are substantially different from those of traditional word analogy datasets. Improving on them is a line of research we wish to investigate further in the future. We make SCAN and the related code publicly available.\footnote{\url{https://github.com/taczin/SCAN_analogies}}

\section{Limitations}
Our experimental design focuses on evaluating the models' analogical capabilities in a generative setting. We see value in this, as analogical reasoning is inherently a generative cognitive function. The BERT models are, however, not trained to perform left-to-right generation and, furthermore, rely on wordpiece vocabulary for tokenization. The evaluation of its predictions in the analogy task is, therefore, less straightforward and not exactly comparable to other models. We adapt the task for BERT models by letting them only predict the first token of the missing answer. Comparing only the first token leaves some variability, however, when matching the prediction and the right answer. We expect this effect to be limited in English due to its sparse morphology. 

Furthermore, the metaphorical analogies come from English literature and cultural background. It would be interesting to compare these with analogies from other languages and cultures to investigate whether the language models' lack of understanding is due to encoding of language-specific properties, missing domain knowledge or the general analogical mapping abilities.

Lastly, some metaphors in SCAN exhibit antiquated gender roles, e.g. the metaphor ``government:household :: governor:father''. While these relationships are culturally often still relevant for metaphor understanding, the underlying implied gender roles need to be treated carefully and issues of their encoding by neural models investigated further. 

\bibliography{anthology,custom}
\bibliographystyle{acl_natbib}

\appendix
\onecolumn
\begin{landscape}
\section{Additional Evaluation Metrics}

\begin{table}[h!]
  \begin{tabular}{|l|l|l|l|l|l|l|l|l|l|l|l|l|l|l|l|l|}
    \hline
    Model &
      \multicolumn{4}{|c|}{BATS} &
      \multicolumn{4}{|c|}{SCAN} &
      \multicolumn{4}{|c|}{SCAN Science} &
      \multicolumn{4}{|c|}{SCAN Meta.}\\
    & Acc & MRR &Rec@10 &Rec@5 & Acc & MRR &Rec@10 &Rec@5 & Acc & MRR &Rec@10 &Rec@5 & Acc & MRR &Rec@10 &Rec@5\\
    \hline
    GloVe &.004 &.099 &.240 &.192 &.018 &.022 &.031 &.024 &.091 &.099 &.110 &.097 &.003 &.006 &.014 &.009 \\
    \hline
    GPT-2 &\textbf{.044} &.098 &\textbf{.250} &\textbf{.172} &.020 &\textbf{.057} &\textbf{.167} &\textbf{.107} &.026 &.073 &.195 &.143 &.019 &\textbf{.054} &\textbf{.161} &\textbf{.099}\\
    \hline
    BERT &\textbf{.126} &\textbf{.207} &\textbf{.401} &\textbf{.316} &.009 &\textbf{.044} &\textbf{.134} &\textbf{.080} &.026 &.092 &\textbf{.273} &.182 &.005 &\textbf{.034} &\textbf{.105} &\textbf{.059}\\
    \hline
    MultBERT &\textbf{.141} &\textbf{.205} &\textbf{.341} &\textbf{.289} &.018 &.041 &\textbf{.100} &\textbf{.067} &.065 &.088 &.130 &.117 &.008 &\textbf{.031} &\textbf{.094} &\textbf{.056}\\
    \hline
  \end{tabular}
  \caption{Accuracy, MRR, Recall@10 and Recall@5 on the two datasets. Statistically significant differences compared to the GloVe baseline are in bold (two-sided permutation test; $p<0.05$; \#resamples=$10e^5$).}
\end{table}

\begin{table}[h!]
  \begin{tabular}{|l|l|l|l|l|l|l|l|l|l|l|l|l|l|l|l|l|}
    \hline
    Model &
      \multicolumn{4}{|c|}{BATS} &
      \multicolumn{4}{|c|}{SCAN} &
      \multicolumn{4}{|c|}{SCAN Science} &
      \multicolumn{4}{|c|}{SCAN Meta.}\\
    & Acc & MRR &Rec@10 &Rec@5 & Acc & MRR &Rec@10 &Rec@5 & Acc & MRR &Rec@10 &Rec@5 & Acc & MRR &Rec@10 &Rec@5\\
    \hline
    GPT-2 &\textbf{.050} &\textbf{.121} &\textbf{.282} &\textbf{.218} &.018 &.048 &.147 &.087 &.013 &.056 &.169 &.104 &.019 &.046 &.142 &.083 \\
    \hline
    BERT &\textbf{.054} &\textbf{.095} &\textbf{.207} &\textbf{.152} &.016 &.035 &.107 &.060 &.052 &.077 &.143 &.117 &.008 &.027 &.099 &.048\\
    \hline
    MultBERT &\textbf{.123} &\textbf{.180} &\textbf{.307} &\textbf{.256} &.020 &.036 &.071 &.051 &.091 &.112 &.143 &.13 &.005 &.020 &.056 &.035\\
    \hline
  \end{tabular}
  \caption{Accuracy, MRR, Recall@10 and Recall@5 when an example sentence is given. Statistically significant differences (two-sided permutation test; $p<0.05$; \#resamples=$10e^5$) compared to each model's baseline in bold.}
\end{table}

\begin{table}[h!]
  \begin{tabular}{|l|l|l|l|l|l|l|l|l|l|l|l|l|l|l|l|l|}
    \hline
    Model &
      \multicolumn{4}{|c|}{BATS} &
      \multicolumn{4}{|c|}{SCAN} &
      \multicolumn{4}{|c|}{SCAN Science} &
      \multicolumn{4}{|c|}{SCAN Meta.}\\
    & Acc & MRR &Rec@10 &Rec@5 & Acc & MRR &Rec@10 &Rec@5 & Acc & MRR &Rec@10 &Rec@5 & Acc & MRR &Rec@10 &Rec@5\\
    \hline
    GPT-2 &\textbf{.305} &\textbf{.384} &\textbf{.550} &\textbf{.482} &.011 &\textbf{.022} &\textbf{.051} &\textbf{.036} &\textbf{.039} &.066 &.143 &.117 &.005 &\textbf{.012} &\textbf{.032} &\textbf{.019}\\
    \hline
    BERT &\textbf{.501} &\textbf{.592} &\textbf{.756} &\textbf{.717} &.004 &\textbf{.019} &\textbf{.051} &\textbf{.031} &.026 &.061 &\textbf{0.130} &.091 &.000 &\textbf{.010} &\textbf{.035} &\textbf{.019}\\
    \hline
    MultBERT &\textbf{.420} &\textbf{.499} &\textbf{.661} &\textbf{.604} &.016 &\textbf{.020} &\textbf{.038} &\textbf{.022} &.078 &.087 &.156 &.078 &.003 &\textbf{.006} &\textbf{.013} &\textbf{.011}\\
    \hline
  \end{tabular}
  \caption{Accuracy, MRR, Recall@10 and Recall@5 on the BATS test set as well as on SCAN after training. Statistically significant differences (two-sided permutation test; $p<0.05$; \#resamples=$10e^5$) compared to each model's baseline in bold.}
\end{table}
\end{landscape}

\end{document}